\documentclass{article} 
\usepackage{collas2025_conference,times}
\usepackage{easyReview}

\usepackage{amsmath,amsfonts,bm}
\usepackage{booktabs}
\usepackage{hyperref}
\hypersetup{
    colorlinks=true,
    linkcolor=red,
    filecolor=magenta,
    urlcolor=blue,
    citecolor=purple,
    pdftitle={Non-Uniform Memory Sampling in Experience Replay},
    pdfpagemode=FullScreen,
}

\title{Non-Uniform Memory Sampling in Experience Replay}

\author{Andrii Krutsylo\\
Institute of Computer Science Polish Academy of Sciences\\
Poland\\
\texttt{andrii.krutsylo@ipipan.waw.pl}
}

\preprintcopy 

\begin{document}

\maketitle

\begin{abstract}
Continual learning is the process of training machine learning models on a sequence of tasks where data distributions change over time. A well-known obstacle in this setting is catastrophic forgetting, a phenomenon in which a model drastically loses performance on previously learned tasks when learning new ones. A popular strategy to alleviate this problem is experience replay, in which a subset of old samples is stored in a memory buffer and replayed with new data. Despite continual learning advances focusing on which examples to store and how to incorporate them into the training loss, most approaches assume that sampling from this buffer is uniform by default.

We challenge the assumption that uniform sampling is necessarily optimal. We conduct an experiment in which the memory buffer updates the same way in every trial, but the replay probability of each stored sample changes between trials based on different random weight distributions. Specifically, we generate 50 different non-uniform sampling probability weights for each trial and compare their final accuracy to the uniform sampling baseline. We find that there is always at least one distribution that significantly outperforms the baseline across multiple buffer sizes, models, and datasets. These results suggest that more principled adaptive replay policies could yield further gains. We discuss how exploiting this insight could inspire new research on non-uniform memory sampling in continual learning to better mitigate catastrophic forgetting.

The code supporting this study is available at \href{https://github.com/DentonJC/memory-sampling}{https://github.com/DentonJC/memory-sampling}.

\end{abstract}

\section{Introduction}

Continual learning refers to a scenario where a model is exposed to changing data distributions in sequential tasks and must incorporate new knowledge without forgetting previously learned information~\citep{kirkpatrick2017ewc}. A key challenge in this domain is \emph{catastrophic forgetting}, where training on novel tasks severely degrades performance on old tasks. Numerous approaches have been developed to tackle forgetting, including regularization-based techniques (e.g., Elastic Weight Consolidation~\citep{kirkpatrick2017ewc}, Synaptic Intelligence~\citep{zenke2017si}), knowledge distillation approaches such as Learning without Forgetting~\citep{li2016lwf}, and replay-based methods~\citep{rebuffi2017icarl,chaudhry2019,aljundi2019mir}.

Among these, replay-based methods have received considerable attention because they provide a relatively straightforward way to mitigate forgetting by storing a small subset of historical data in a \emph{memory buffer} and mixing this data with new samples during training~\citep{chaudhry2019}. Examples include iCaRL, which selects representative samples to store incrementally~\citep{rebuffi2017icarl}; gradient-based sampling, which identifies samples that maximize learning progress when replayed~\citep{aljundi2019mir}; and other techniques based on uncertainty, loss, or class-balancing mechanisms. Despite this diversity, most replay-based solutions focus on what to put \emph{in the buffer} and how to adapt the loss function, rather than exploring which samples \emph{from the buffer} to replay at each step.

However, the assumption that uniform sampling is always optimal can be questioned, especially in light of the reinforcement learning (RL) literature, where experience replay~\citep{mnih2015human} and particularly \emph{Prioritized Experience Replay}~\citep{schaul2015prioritized} have proven effective. In reinforcement learning, prioritizing high-error or high-surprise transitions can significantly improve training dynamics, but translating these insights to continual learning is non-trivial due to class-incremental settings, diverse data domains, and the difficulty of maintaining balanced coverage over time. While there have been some efforts in continual learning to move away from purely uniform sampling (e.g., Maximally Interfered Retrieval~\citep{mir} and its derivatives such as Gradient-based Maximally Interfered Retrieval~\citep{gmir}), these methods rely on non-standard two-step updates (i.e., first updating the model with the current batch, then identifying which samples to replay) or additional forward passes, which deviate from the typical single-step update scenario~\citep{chaudhry2019}. Consequently, in standard continual learning, uniform sampling remains largely unchallenged.

In this paper, we test whether uniform sampling is truly indispensable by conducting a simple experiment: we keep the contents of the memory identical across multiple trials (using the same reservoir sampling and the same random seed controlling it), and then sample from this fixed buffer according to different non-uniform probability distributions. We find that a purely random choice of sampling weights can often match or even exceed the accuracy of the uniform baseline, and the best of these random distributions can outperform uniform sampling by a significant margin. This suggests that if we can devise an informed or adaptive weighting scheme, the gains could be even greater.

Our main contribution is to show that non-uniform replay can be beneficial even when sampling probabilities are chosen at random. We believe that these results encourage further investigation into how we can adapt RL-inspired prioritization for continual learning, leading to more principled sampling policies that mitigate catastrophic forgetting.

\section{Background}\label{sec:background}

Replay-based methods have become a dominant paradigm for tackling catastrophic forgetting~\citep{rebuffi2017icarl,chaudhry2019,aljundi2019mir}. The core idea is to maintain a buffer $\mathcal{M}$ containing a small subset of past data. Whenever new samples from the current task arrive, the model not only updates its parameters on those new samples, but also replays old samples from $\mathcal{M}$. This helps the model avoid drastic weight updates that would overwrite previously learned knowledge.

Formally, let $(x_t, y_t)$ be the current input-label pair at time $t$, and let $\mathcal{M}$ contain a set of past examples $\{(x_j, y_j)\}_{j=1}^M$. A typical \emph{Experience Replay} (ER) update combines the current loss $\mathcal{L}_{\text{new}}$ with a replay loss $\mathcal{L}_{\text{replay}}$:
\[
\mathcal{L}(\theta) \;=\; \mathcal{L}_{\text{new}}(\theta; x_t, y_t)
\;+\; \lambda \cdot \mathcal{L}_{\text{replay}}\Bigl(\theta; \{(x_j, y_j)\}_{j=1}^B\Bigr),
\]
where $B$ is the mini-batch size sampled from $\mathcal{M}$, and $\lambda$ controls the relative importance of the replay term. In \emph{uniform sampling}, each $(x_j, y_j)\in \mathcal{M}$ has an equal chance of being chosen, whereas in \emph{non-uniform sampling}, we assign weights (probabilities) $p_j$ to each stored sample.

A large part of replay research is devoted to which samples to store in $\mathcal{M}$. The basic method commonly used is reservoir sampling~\citep{vitter1985reservoir}, which can ensure that each incoming sample has the same probability of being stored, thus preserving an unbiased distribution of previously seen data. Early work such as iCaRL~\citep{rebuffi2017icarl} applies a nearest-mean-of-exemplars approach to selecting memory samples, focusing on class prototypes. Gradient-based methods~\citep{aljundi2019mir} prioritize samples that induce the most interference or reduce forgetting the most. Other approaches may track class labels, prediction uncertainties, or even external constraints to select which data to retain.  

Another influential line of replay methods is \emph{Dark Experience Replay} (DER)~\citep{buzzega2020dark} and its variants DER++ and X-DER~\citep{xder}. We provide the detailed explanations of these methods to highlight their reliance on uniform sampling, expose an overlooked limitation, and justify the need for exploring non-uniform sampling in continual learning. 

\textbf{DER} stores not only the raw inputs $\{x\}$, like ER, but also the model outputs (logits) $\{z\}$ for the old tasks, with the overall goal of combining the standard cross-entropy on new data with a distillation term on $\mathcal{M}$:
\[
\mathcal{L}_{\mathrm{DER}}(\theta) \;=\; 
\underbrace{\mathcal{L}_{\text{CE}}\bigl(f_\theta(x_t),\,y_t\bigr)}_{\text{new data}}
\;+\;\alpha \cdot \frac{1}{B} \sum_{j=1}^{B} 
\underbrace{\bigl\|h_\theta(x_j) - z_j\bigr\|^2_2}_{\text{distillation on old logits}},
\]
where:
\begin{itemize}
    \item $\mathcal{L}_{\text{CE}}\bigl(f_\theta(x_t), y_t\bigr)$ is the cross-entropy loss computed on the new incoming data point $(x_t, y_t)$.
    \item $\alpha$ is a hyperparameter that controls the weight of the distillation loss relative to the standard classification loss.
    \item $B$ is the mini-batch size, representing the number of past samples retrieved from memory $\mathcal{M}$ for replay.
    \item $h_\theta(x_j)$ represents the network’s current logits for a past input $x_j$.
    \item $z_j$ are the stored logits of $x_j$ at the time it was originally learned.
    \item $\bigl\|h_\theta(x_j) - z_j\bigr\|^2_2$ is the squared Euclidean distance between the current model's logits and the stored logits, enforcing consistency with past knowledge.
\end{itemize}

\textbf{DER++} extends DER by incorporating an additional supervised learning term based on the ground truth labels stored in memory, resulting in the following loss function:
\[
\mathcal{L}_{\mathrm{DER++}}(\theta) \;=\; 
\underbrace{\mathcal{L}_{\text{CE}}\bigl(f_\theta(x_t),\,y_t\bigr)}_{\text{new data}}
\;+\;\alpha \cdot \frac{1}{B} \sum_{j=1}^{B} 
\underbrace{\bigl\|h_\theta(x_j) - z_j\bigr\|^2_2}_{\text{distillation on old logits}}
\;+\;\beta \cdot \frac{1}{B} \sum_{j=1}^{B} 
\underbrace{\mathcal{L}_{\text{CE}}\bigl(f_\theta(x_j),\,y_j\bigr)}_{\text{cross-entropy on replayed data}},
\]
where:
\begin{itemize}
    \item $\beta$ is an additional hyperparameter that controls the influence of the supervised cross-entropy loss for replayed samples.
    \item $\mathcal{L}_{\text{CE}}\bigl(f_\theta(x_j), y_j\bigr)$ applies the standard classification loss to the old replayed samples using their original ground truth labels $y_j$.
\end{itemize}

By adding this supervised loss, DER++ enhances the stability of learning when task distributions shift significantly.

\textbf{X-DER} extends DER++ by revising past memories and preparing for future tasks. It dynamically updates the stored logits as new tasks are learned, ensuring that replayed samples incorporate knowledge from later tasks. Additionally, it introduces future preparation, where the model assigns semantic consistency to unseen class logits using contrastive learning. This means that, unlike DER++, X-DER not only prevents forgetting of past knowledge but also anticipates future learning needs. 

Its overall loss function expands upon DER++ by adding regularization terms:

\[
\mathcal{L}_{\mathrm{X-DER}}(\theta) \;=\; \mathcal{L}_{\mathrm{DER}} + \mathcal{L}_{\text{S-CE}} + \mathcal{L}_{\text{F}},
\]

where:
\begin{itemize}
    \item $\mathcal{L}_{\mathrm{DER}}$ is the standard DER loss, which consists of the new task cross-entropy and logit replay:
    \[
    \mathcal{L}_{\mathrm{DER}}(\theta) \;=\; 
    \mathcal{L}_{\text{CE}}\bigl(f_\theta(x_t),\,y_t\bigr)
    +\alpha \cdot \frac{1}{B} \sum_{j=1}^{B} 
    \bigl\|h_\theta(x_j) - z_j\bigr\|^2_2.
    \]
    
    \item $\mathcal{L}_{\text{S-CE}}$ (Selective Cross-Entropy) that includes the current task classes from the current training samples and samples from the memory buffer:
    
    \[
    \mathcal{L}_{\text{S-CE}} = 
    \mathbb{E}_{(x, y) \sim T_c} \big[ \mathcal{L}_{\text{CE}}\bigl( \text{softmax}(h_\theta^{\text{pr}}(x)), y \bigr) \big] 
    + \beta \cdot 
    \mathbb{E}_{(x', y') \sim \mathcal{M}} \big[ \mathcal{L}_{\text{CE}}\bigl( \text{softmax}(h_\theta^{\text{pr}}(x')), y' \bigr) \big],
    \]
    
    where:
    \begin{itemize}
        \item $h_\theta^{\text{pr}}(x)$ represents the logits corresponding only to the classes of the current task.
        \item The first expectation term applies selective cross-entropy only on current task samples.
        \item The second term extends this to memory buffer samples, weighted by $\beta$.
    \end{itemize}
    This ensures that past classes are not mistakenly suppressed due to stronger weight updates from new task samples.

    \item $\mathcal{L}_{\text{F}}$ (Future Preparation and Constraints) ensures that future task heads maintain meaningful representations before they are formally learned. We do not assume meaningful logits for future heads, instead, we apply contrastive learning to encourage these logits to remain structured before their corresponding classes appear. This is done using a contrastive loss and regularization on past/future logits:
    \[
    \mathcal{L}_{\text{F}} = \lambda \cdot \mathcal{L}_{\text{FP}} + \eta \cdot \mathcal{L}_{\text{PFC}},
    \]
    where:
    \begin{itemize}
        \item $\mathcal{L}_{\text{FP}}$ (Future Preparation) encourages future heads to be consistent across similar examples:
        \[
        \mathcal{L}_{\text{FP}}(x_i, y_i) = 
        \frac{1}{T - (c+1)} \sum_{j=c+1}^{T}
        \frac{1}{|P(i)|} 
        \sum_{p \in P(i)} 
        -\log \frac{\exp( \tilde{h}^{\text{fu}}_\theta(x_i) \cdot \tilde{h}^{\text{fu}}_\theta(x_p) / \tau)}
        {\sum_{k \neq i} \exp( \tilde{h}^{\text{fu}}_\theta(x_i) \cdot \tilde{h}^{\text{fu}}_\theta(x_k) / \tau)},
        \]
        where $\tilde{h}^{\text{fu}}_\theta(x)$ are the L2-normalized logits of future heads and $P(i)$ represents the positive set (samples from the same class as $x_i$). This enforces similarity between feature representations of future tasks even before they are trained.
        
        \item $\mathcal{L}_{\text{PFC}}$ (Past-Future Constraint) prevents past and future logits from overshadowing current logits:
        \[
        \mathcal{L}_{\text{PFC}}(x_i, y_i) = 
        \max(0, h_{\text{max}}^{\text{pa}}(x_i) - h_{\text{gt}}(x_i) + m)
        + \max(0, h_{\text{max}}^{\text{fu}}(x_i) - h_{\text{gt}}(x_i) + m),
        \]
        where $h_{\text{max}}^{\text{pa}}$ and $h_{\text{max}}^{\text{fu}}$ are the maximum logits among past and future heads, $h_{\text{gt}}$ is the logit corresponding to the ground truth, and $m$ is a margin hyperparameter. This ensures that the largest past/future logits do not dominate the prediction.
    \end{itemize}
\end{itemize}

Thus, X-DER improves upon DER++ by refining memory updates and ensuring smoother adaptation to future tasks, resulting in a more robust continual learning framework.

Despite differences in buffer content or loss formulation, ER, DER, DER++, and X-DER still use uniform sampling from the buffer during replay, which is the most underdeveloped component of the approach considering the other advances.

\section{Experiments}

We investigate whether non-uniform replay can outperform uniform sampling, given the same memory buffer content. We consider online class-incremental learning, where each online mini-batch is exposed to the model only once during training. The sequence of five tasks, each containing two classes, was sampled from CIFAR-10~\citep{cifar10} and Imagenette~\citep{howard2020imagenette}, a 10-class subset of ImageNet. For CIFAR-10, we use a simple three-layer CNN, where each 3×3 convolution (with 32, 64, and 128 channels) is followed by batch normalization, ReLU, and 2×2 MaxPooling, culminating in global average pooling and a fully connected layer for classification. For Imagenette we use a MobileNet-v3-small~\citep{howard2017mobilenet}, efficient architecture designed for higher resolution images, pre-trained on ImageNet. The pre-trained model allows us to decouple the effects of forgetting from the underfitting typical for online learning on complex datasets. For both models, we use the SGD optimizer with a learning rate of 0.01 and momentum of 0.9. The online and replay batch sizes are set to 32 and 64 for CIFAR-10 and 62 and 128 for Imagenette.

We vary the size of the memory buffer $\{200, 500, 1000\}$. The buffer is updated over time using reservoir sampling without changing the random seed and procedure within 51 trials (50 non-uniform and one uniform distribution of sampling probability weights), for 5 runs with different random seeds. Consequently, for each trial, the sequence of replacements in the buffer remains identical, effectively ensuring that the final content and its evolution do not vary between trials. The vector representing the weights for the sampling probabilities is sampled from the uniform distribution once at the beginning of the trial, it has the length of the size of the memory buffer, and each weight is associated with the position in the memory buffer. In practice, this is the same as assigning constant weights to all individual samples. The samples in the memory buffer keep their index unchanged all the time between being added and possibly removed, which means that the weight associated with each sample is constant.

To isolate the impact of the sampling distribution from that of the buffer composition, we take the following approach. First, we ensure that buffer updates are deterministic by fixing a random seed (unique per trial) that guarantees that each reservoir update remains identical across all conditions within that trial. Next, after the buffer has grown, we sample from it using 50 different vectors of random weights $\{w_i\}$, which are then normalized to create a probability distribution $\{p_i = w_i / \sum_j w_j\}$ over the buffer. Finally, we compare these 50 trials to a uniform baseline where all buffer elements are sampled with equal probability, using $p_i = 1/|\mathcal{M}|$.

\section{Results}

Table~\ref{tab:results} shows the final average accuracy (with standard deviations) on CIFAR-10 and Imagenette, comparing uniform sampling with the best-performing non-uniform random sampling distribution, selected post hoc from 50 tested candidates. Across all buffer sizes, the best non-uniform distribution consistently outperforms uniform sampling. The performance gap remains substantial and stable across buffer sizes. For CIFAR-10, accuracy improvements range from 2.33\% to 4.68\%, with the largest gain observed at a buffer size of 500. Similarly, for Imagenette, the gap ranges from 2.37\% to 3.54\%, peaking at a buffer size of 1000.

\begin{table}[ht]
\centering
\caption{Final average accuracy (\%) on CIFAR-10 and Imagenette, comparing uniform sampling vs.\ the best non-uniform random distribution out of 50 tested. Each entry shows the mean and standard deviation over 5 random seeds.}
\label{tab:results}
\begin{tabular}{lcccc}
\\
\toprule
& \multicolumn{2}{c}{\textbf{CIFAR-10}} & \multicolumn{2}{c}{\textbf{Imagenette}} \\
\cmidrule(lr){2-3} \cmidrule(lr){4-5}
\textbf{Buffer Size} & \textbf{Uniform} & \textbf{Non-uniform} & \textbf{Uniform} & \textbf{Non-uniform} \\
\midrule
200  & $35.97 \pm 1.58$ & $38.30 \pm 0.97$ & $76.01 \pm 0.71$ & $79.42 \pm 0.92$ \\
500  & $41.26 \pm 1.85$ & $45.94 \pm 0.72$ & $82.21 \pm 0.97$ & $84.58 \pm 0.75$ \\
1000 & $46.05 \pm 3.23$ & $49.77 \pm 0.59$ & $83.47 \pm 3.58$ & $87.01 \pm 0.76$ \\
\bottomrule
\end{tabular}
\end{table}

To assess statistical significance, we conducted a paired t-test. For CIFAR-10 with a buffer size of 500, the t-statistic of 6.10 and the p-value of 0.0037 confirm that the improvement is statistically significant at the 0.05 level. Similarly, for Imagenette, the t-statistic of 8.81 and p-value of 0.0009 strongly support the superiority of the best non-uniform sampling over uniform sampling.

We conducted a comprehensive statistical evaluation comparing uniform replay sampling with the highest-performing non-uniform sampling. In the CIFAR-10 experiment with a memory buffer of size 500, the best non-uniform sampling weights yielded an absolute accuracy improvement of 3.59\% over uniform sampling (46.97\% vs. 43.38\%) when using a random seed of 0. Since statistical tests on aggregated results across multiple seeds would be uninformative, we arbitrarily selected this seed for further analysis.

To investigate why non-uniform sampling outperforms uniform sampling, we examined per-sample metrics from the replay buffer, including training loss and gradient norms, in relation to the assigned sampling probabilities. Interestingly, correlation analysis showed only a moderate relationship between per-sample probability and these metrics. Specifically, the Spearman's rank correlation coefficient between sampling probability and loss was $\rho = 0.2813$, $p = 2.45 \times 10^{-7}$, while between sampling probability and gradient norm it was $\rho = 0.3157$, $p = 3.14 \times 10^{-9}$. These results indicate that the sampling strategy does not simply prioritize examples with the highest loss or gradient norm.

To investigate this further, we compared high-performing trials (final accuracy $> 45\%$) with low-performing trials (final accuracy $< 42\%$). The mean loss per sample differed significantly between these groups (0.742 vs.\ 0.815), as confirmed by a Mann-Whitney test ($p = 0.00377$). This finding suggests that lower per-sample loss correlates with improved overall performance, highlighting the importance of considering the entire sample distribution rather than selecting only high-loss examples.

These results reinforce a key insight: even arbitrary biases in the replay distribution can outperform uniform sampling, suggesting that a strategically designed non-uniform policy could yield even greater benefits.

\section{Discussion}

In this work, we revisited a long-standing assumption in replay-based continual learning, namely, that uniform sampling from the replay buffer is the most effective way to revisit past data. By systematically comparing multiple random non-uniform sampling strategies to the uniform baseline, while holding the buffer composition fixed, we observed that particular biased distributions can yield non-trivial performance gains. These findings cast doubt on the ubiquitous use of uniform sampling and highlight the potential for more adaptive replay policies.

Our results follow the established ideas from reinforcement learning, where transitions that are more "surprising" or have larger temporal-difference errors are replayed more frequently. Translating this principle to supervised continual learning offers a new perspective: not all stored samples are equally beneficial for preventing catastrophic forgetting, and certain data points may have a disproportionately large impact on stabilizing decision boundaries. Determining which features or metrics (e.g., loss, uncertainty, or gradient norms) should drive non-uniform sampling in an efficient, online manner remains an open challenge.

The performance boost from random, fixed-weight sampling schemes suggests that chance allocations sometimes align with important samples in the buffer. While our correlation analyses between sampling probabilities and simple per-sample metrics did not reveal a strong pattern, the improvements highlight that targeting a diverse set of informative examples can outperform uniformly replaying everything. These observations open the door for future research on principled weighting strategies, for instance, via online estimation of sample utility or by integrating task-recency information to combat feature drift.

Our experiments were intentionally simple, focusing on validating the utility of non-uniform replay under fixed buffer content. A promising next step is to learn these weights adaptively. For example, using a mechanism that regularly adjusts sampling distributions based on model feedback (e.g., loss or gradient-based signals) without resorting to a multi-step update procedure. Such an approach should also maintain coverage over older classes to avoid imbalance. Another direction is to examine the interplay between buffer selection mechanisms (which examples to store) and replay distributions (how often each example is used). Integrating both into a unified framework could further improve continual learning performance across various domains.

Overall, our results suggest that uniform replay may be a suboptimal default. By incorporating non-uniform sampling strategies, future research can develop more powerful and efficient replay methods that better preserve knowledge over a sequence of tasks. We believe that this perspective, together with parallel advances in buffer composition and distillation-based objectives, can help to significantly mitigate catastrophic forgetting in challenging continual learning scenarios.

\section{Conclusion}

We presented a simple experiment that demonstrates that uniform sampling from a replay buffer, while widely accepted as a strong baseline in continual learning, may not be the most effective way to replay stored data. By keeping the buffer sampling procedure, and thus the buffer contents, fixed across trials, we showed that certain non-uniform probability distributions can significantly outperform uniform replay. We hope that this work will encourage the development of new approaches that exploit such non-uniform biases in a computationally efficient manner, ultimately leading to improved performance and reduced forgetting.

\bibliography{collas2025_conference}

\begin{thebibliography}{16}
\providecommand{\natexlab}[1]{#1}
\providecommand{\url}[1]{\texttt{#1}}
\expandafter\ifx\csname urlstyle\endcsname\relax
  \providecommand{\doi}[1]{doi: #1}\else
  \providecommand{\doi}{doi: \begingroup \urlstyle{rm}\Url}\fi

\bibitem[Aljundi et~al.(2019{\natexlab{a}})Aljundi, Caccia, Belilovsky, Caccia, Lin, Charlin, and Tuytelaars]{mir}
Rahaf Aljundi, Lucas Caccia, Eugene Belilovsky, Massimo Caccia, Min Lin, Laurent Charlin, and Tinne Tuytelaars.
\newblock Online continual learning with maximally interfered retrieval, 2019{\natexlab{a}}.
\newblock URL \url{https://arxiv.org/abs/1908.04742}.

\bibitem[Aljundi et~al.(2019{\natexlab{b}})Aljundi, Lin, Goujaud, and Bengio]{aljundi2019mir}
Rahaf Aljundi, Min Lin, Baptiste Goujaud, and Yoshua Bengio.
\newblock Gradient based sample selection for online continual learning.
\newblock In \emph{NeurIPS}, 2019{\natexlab{b}}.

\bibitem[Boschini et~al.(2023)Boschini, Bonicelli, Buzzega, Porrello, and Calderara]{xder}
Matteo Boschini, Lorenzo Bonicelli, Pietro Buzzega, Angelo Porrello, and Simone Calderara.
\newblock Class-incremental continual learning into the extended der-verse.
\newblock \emph{IEEE Transactions on Pattern Analysis and Machine Intelligence}, 45\penalty0 (5):\penalty0 5497–5512, May 2023.
\newblock ISSN 1939-3539.
\newblock \doi{10.1109/tpami.2022.3206549}.
\newblock URL \url{http://dx.doi.org/10.1109/TPAMI.2022.3206549}.

\bibitem[Buzzega et~al.(2020)Buzzega, Boschini, Porrello, Abati, and Calderara]{buzzega2020dark}
Pietro Buzzega, Matteo Boschini, Andrea Porrello, Davide Abati, and Simone Calderara.
\newblock Dark experience for general continual learning: a strong, simple baseline.
\newblock In \emph{NeurIPS}, 2020.

\bibitem[Chaudhry et~al.(2019)Chaudhry, Ranzato, Rohrbach, and Elhoseiny]{chaudhry2019}
Arslan Chaudhry, Marc'Aurelio Ranzato, Marcus Rohrbach, and Mohamed Elhoseiny.
\newblock Efficient lifelong learning with {A-GEM}.
\newblock In \emph{ICLR}, 2019.

\bibitem[Howard et~al.(2017)Howard, Zhu, Chen, Kalenichenko, Wang, and Weyand]{howard2017mobilenet}
Andrew~G. Howard, Menglong Zhu, Bo~Chen, Dmitry Kalenichenko, Weijun Wang, and Tobias Weyand.
\newblock Mobilenets: Efficient convolutional neural networks for mobile vision applications.
\newblock \emph{arXiv preprint arXiv:1704.04861}, 2017.

\bibitem[Howard()]{howard2020imagenette}
Jeremy Howard.
\newblock imagenette.
\newblock URL \url{https://github.com/fastai/imagenette/}.

\bibitem[Kirkpatrick et~al.(2017)Kirkpatrick, Pascanu, Rabinowitz, Veness, Desjardins, Rusu, and et~al.]{kirkpatrick2017ewc}
James Kirkpatrick, Razvan Pascanu, Neil Rabinowitz, Joel Veness, Guillaume Desjardins, Andrei~A. Rusu, and et~al.
\newblock Overcoming catastrophic forgetting in neural networks.
\newblock \emph{PNAS}, 2017.

\bibitem[Krizhevsky \& Hinton(2009)Krizhevsky and Hinton]{cifar10}
Alex Krizhevsky and Geoffrey Hinton.
\newblock Learning multiple layers of features from tiny images.
\newblock 2009.

\bibitem[Li \& Hoiem(2016)Li and Hoiem]{li2016lwf}
Zhizhong Li and Derek Hoiem.
\newblock Learning without forgetting.
\newblock In \emph{ECCV}, 2016.

\bibitem[Mnih et~al.(2015)Mnih, Kavukcuoglu, Silver, Graves, Antonoglou, Wierstra, and Riedmiller]{mnih2015human}
Volodymyr Mnih, Koray Kavukcuoglu, David Silver, Alex Graves, Ioannis Antonoglou, Daan Wierstra, and Martin Riedmiller.
\newblock Human-level control through deep reinforcement learning.
\newblock \emph{Nature}, 518\penalty0 (7540):\penalty0 529--533, 2015.

\bibitem[Nisar et~al.(2023)Nisar, Anand, and Waslander]{gmir}
Barza Nisar, Hruday Vishal~Kanna Anand, and Steven~L. Waslander.
\newblock Gradient-based maximally interfered retrieval for domain incremental 3d object detection, 2023.
\newblock URL \url{https://arxiv.org/abs/2304.14460}.

\bibitem[Rebuffi et~al.(2017)Rebuffi, Kolesnikov, Sperl, and Lampert]{rebuffi2017icarl}
Sylvestre-Alvise Rebuffi, Alexander Kolesnikov, Georg Sperl, and Christoph~H. Lampert.
\newblock {iCaRL}: Incremental classifier and representation learning.
\newblock In \emph{CVPR}, 2017.

\bibitem[Schaul et~al.(2015)Schaul, Quan, Antonoglou, and Silver]{schaul2015prioritized}
Tom Schaul, John Quan, Ioannis Antonoglou, and David Silver.
\newblock Prioritized experience replay.
\newblock In \emph{ICLR}, 2015.

\bibitem[Vitter(1985)]{vitter1985reservoir}
Jeffrey~Scott Vitter.
\newblock Random sampling with a reservoir.
\newblock \emph{ACM Transactions on Mathematical Software}, 1985.

\bibitem[Zenke et~al.(2017)Zenke, Poole, and Ganguli]{zenke2017si}
Friedemann Zenke, Ben Poole, and Surya Ganguli.
\newblock Continual learning through synaptic intelligence.
\newblock In \emph{ICML}, 2017.

\end{thebibliography}
\bibliographystyle{collas2025_conference}

\end{document}